\documentclass[letterpaper, 10pt, conference]{ieeeconf}

\IEEEoverridecommandlockouts
\usepackage{graphicx}
\usepackage{xcolor}
\usepackage{booktabs}
\usepackage{array}
\usepackage{amsmath}
\usepackage{amssymb}
\usepackage{tikz}
\usetikzlibrary{shapes,arrows,positioning,fit,calc}
\usepackage{listings}
\usepackage{algorithm}
\usepackage{algorithmic}
\usepackage{hyperref}
\hypersetup{
    colorlinks=false,
    linkcolor=black,
    citecolor=black,
    urlcolor=black,
    pdfauthor={Hamied Nabizada, Thomas Wirt, Luis Miguel Vieira da Silva, Felix Gehlhoff, Alexander Fay},
    pdftitle={From Capability Models to Automated Planning: An AAS-Native Approach for Automatic PDDL Generation},
    pdfkeywords={Asset Administration Shell, AAS, PDDL, automated planning, capability models, Industry 4.0, digital twin, flexible manufacturing, model-driven engineering}
}

\lstdefinelanguage{PDDL}{
    keywords={define, domain, problem, requirements, types, constants, predicates,
              functions, action, parameters, precondition, effect, objects, init,
              goal, and, or, not, when, forall, exists, either, imply, increase,
              decrease, assign, at, over, all, start, end, duration, durative-action,
              condition, :strips, :typing, :equality, :fluents, :durative-actions,
              :action-costs, :negative-preconditions, :disjunctive-preconditions,
              :conditional-effects, :adl},
    sensitive=true,
    morecomment=[l]{;},
    morestring=[b]",
    alsoletter={:,-},
}

\lstdefinestyle{pddl}{
    language=PDDL,
    basicstyle=\ttfamily\footnotesize,
    keywordstyle=\color{blue}\bfseries,
    commentstyle=\color{gray}\itshape,
    stringstyle=\color{red},
    showstringspaces=false,
    numbers=left,
    numberstyle=\tiny\color{gray},
    numbersep=5pt,
    frame=single,
    breaklines=true,
    breakatwhitespace=true,
    tabsize=2,
    captionpos=b,
}

\title{\LARGE \bf
From Capability Models to Automated Planning: \\
An AAS-Native Approach for Automatic PDDL Generation
}

\author{
    Hamied Nabizada$^{1}$, Thomas Wirt$^{1}$, Luis Miguel Vieira da Silva$^{1}$, Felix Gehlhoff$^{1}$, and Alexander Fay$^{2}$
    \thanks{This contribution originates from the projects \textit{iMOD} and \textit{RIVA}, funded by \textit{dtec.bw -- Digitalization and Technology Research Center of the Bundeswehr}, which we gratefully acknowledge. \textit{dtec.bw} is funded by the \textit{European Union -- NextGenerationEU}.}%
    \thanks{$^{1}$Institute of Automation Technology, Helmut Schmidt University Hamburg, Hamburg, Germany.}%
    \thanks{$^{2}$Chair of Automation Technology, Ruhr University Bochum, Bochum, Germany.}
}

\begin{document}

\maketitle
\thispagestyle{empty}
\pagestyle{empty}

\begin{abstract}
Engineers designing production systems need to verify that a given layout supports all required production sequences. 
Automated planning techniques can answer such questions, but formulating the required planning problems in the Planning Domain Definition Language (PDDL) demands specialized expertise that production engineers typically lack.
Asset Administration Shells (AAS) have emerged as the standardized Digital Twin for industrial assets in Industry~4.0.
We show that AAS capability models, structured using four established Industry 4.0 standards~(VDI3682 for process descriptions, IEC61360-1 for semantic property qualification, IDTA02011 for type hierarchies, and IDTA02016 for instance descriptions), contain sufficient information to generate complete PDDL problems automatically.  
Unlike prior work that introduced PDDL-specific submodels, our approach derives all planning elements from domain-level descriptions of resource functions, so-called capabilities, allowing engineers to model capabilities without any exposure to PDDL syntax or planning concepts.
Our extraction algorithm transforms distributed Multi-AAS architectures into complete PDDL planning problems. 
We validate the approach on AAS models of a laboratory production system, comparing four layout variants using optimal planning to demonstrate how engineers can systematically explore design trade-offs by modifying the AAS model and regenerating the planning domain. 
\end{abstract}

\section{Introduction}
\label{sec:introduction}

When designing production systems, engineers must evaluate whether a proposed layout supports all required production sequences, whether additional equipment is necessary, and what the minimal viable configuration looks like~\cite{nabizada2024workflow}.
Answering such questions early in the design process can prevent costly modifications during commissioning or operation~\cite{koecher2023agenda}.
Automated planning techniques, particularly those based on the Planning Domain Definition Language~(PDDL)~\cite{mcdermott1998pddl}, can address such questions~\cite{kootbally2015pddl}.
Given a formal domain and problem specification, automated planners can determine whether goals are achievable and compute action sequences to reach them. 
This enables engineers to explore different production system designs early in the development process (design space exploration)~\cite{nabizada2025keps}. 
However, creating PDDL models requires specialized expertise~\cite{lindsay2023pddl,mayrdorn2022pddl} that production engineers typically lack. 

At the same time, Industry~4.0 initiatives promote the use of Asset Administration Shells (AAS) as standardized Digital Twins for manufacturing equipment~\cite{idta2023aas,novak2022digitalized}. 
In Cyber-Physical Production Systems, such standardized representations enable interoperability across heterogeneous equipment~\cite{monostori2014cpps}. 
Recent works have proposed formal capability models for manufacturing, describing resource functions in an implementation-independent way~\cite{koecher2020capability}, and explored their mapping between ontologies and AAS~\cite{vieira2023mapping}.
These capability models originate from the automation engineering domain, whereas PDDL semantics typically do not, and thus are not natively supported or integrated into existing models.

However, since AAS models already capture structured descriptions of production resources and their functions, the question arises whether PDDL planning problems can be derived directly from these models, without additional PDDL-specific modeling effort. %The question arises whether PDDL planning problems can be derived directly from AAS capability models, without additional PDDL-specific modeling effort. 
Building on work that demonstrated semantic alignment between generic capability models, based on standards such as VDI3682 and IEC61360-1, and PDDL~\cite{vieira2023pddl}, we present an AAS-native approach that transforms standardized capability models into complete PDDL planning problems. %Building on prior work that demonstrated semantic alignment between generic capability md, based on standards such as VDI3682 and IEC61360-1, and PDDL~\cite{vieira2023pddl}, we present an AAS-native approach that transforms standardized capability models into complete PDDL planning problems. 
Unlike approaches that introduce PDDL-specific AAS submodels~\cite{bernhard2024aas}, our method derives planning information from established capability standards and can reuse existing capability models without requiring PDDL knowledge. 

We address two research questions:
\begin{enumerate}
\item How can AAS capability models be systematically transformed into PDDL planning problems?
\item How can this transformation handle distributed Multi-AAS architectures where planning information is spread across multiple files?
\end{enumerate}
We validate our approach on a laboratory production system, demonstrating design space exploration across four layout variants.

The remainder of this paper is structured as follows: Section~\ref{sec:background} provides background on AAS, PDDL, and relevant standards, and discusses related work. 
Section~\ref{sec:approach} presents our approach including the mapping, the algorithm, and the implementation. 
Section~\ref{sec:evaluation} presents a case study.
Section~\ref{sec:conclusion} concludes the paper, discusses limitations, and provides an outlook.

\section{Background and Related Work}
\label{sec:background}

\subsection{Asset Administration Shell}

The AAS is the standardized Digital Twin implementation for Industry~4.0, defined by \emph{Plattform Industrie~4.0}~\cite{plattformi40_2022} and specified by the Industrial Digital Twin Association (IDTA)~\cite{idta2023aas}.
An AAS provides a structured, machine-readable description of an asset's properties, capabilities, and relationships.
AAS packages are serialized as AASX files, a standardized container format. 

The AAS metamodel organizes information into \emph{submodels}, each grouping a specific aspect of an asset such as technical data, nameplate information, or capabilities.
The IDTA publishes standardized \emph{submodel templates} that define structure and semantics for particular domains; these templates are continuously extended as new application areas emerge.
Submodels contain hierarchically organized elements: \emph{SubmodelElementCollections} (SMC) that group related elements, \emph{Properties} that store typed values, and \emph{ReferenceElements} that enable cross-references between submodels and across different AAS files.
This referencing mechanism is central to our approach, as it supports modular system composition where each component is described in its own AAS file.

Two of these submodel templates are particularly relevant for our approach: IDTA02011~\cite{idta2023hierarchical} defines hierarchical structures through Entity elements with parent-child relationships, enabling type hierarchies within a submodel. 
IDTA02016~\cite{idta2023instance} provides a template for modeling concrete component instances with their states. 
Section~\ref{sec:approach} details how these elements map to PDDL constructs. 

\subsection{PDDL and Automated Planning}

PDDL~\cite{mcdermott1998pddl} is the de facto standard input language for automated planning, which generates plans by automatically computing action sequences that transform an initial state into a desired goal state. 
A PDDL specification consists of a \emph{domain} file defining the possible actions that change the state of objects, and a \emph{problem} file defining a specific scenario to solve~\cite{haslum2019pddl}. 

The domain file specifies: \textbf{Types} (a hierarchy of object categories), \textbf{Predicates} (boolean properties with typed parameters), and \textbf{Actions} (state transitions with preconditions and effects). 
Each action defines \emph{parameters} (typed variables), \emph{preconditions} (predicates that must hold), and \emph{effects} (predicates added or removed). 
The problem file specifies: \textbf{Objects} (typed instances), the \textbf{Initial State} (facts that hold initially), and the \textbf{Goal} (conditions to achieve). 
Our approach generates PDDL domain and problem files covering the elements shown in Table~\ref{tab:mapping}.

A planner~(a solver for automated planning problems) takes domain and problem files as input and computes a sequence of ground actions (a \emph{plan}) that transforms the initial state into one satisfying the goal.
A key advantage of PDDL as an interchange format is \emph{planner and domain independence}: the language can express planning problems from any application domain, since any standard-compliant planner can process the generated files, and users can choose between different planning strategies without modifying the model.
\emph{Satisficing} planners find valid plans quickly, while \emph{optimal} planners guarantee the shortest plan at higher computational cost.
Abstraction frameworks such as the Unified Planning Framework (UPF)~\cite{upf2025} further simplify planner integration by providing a common API across multiple planning engines.
Since our approach generates standard PDDL files, any compliant planner can be used; the choice of the planning strategy is independent of the generation process.

\subsection{Capability Modeling in Manufacturing}
Capability-based approaches describe manufacturing resources in terms of \emph{what they can do} rather than \emph{how they are built}~\cite{perzylo2019capability,jarvenpaa2016formal}. 
K\"ocher et al.~\cite{koecher2023reference} consolidated this research into the Capability-Skill-Service (CSS) reference model, developed jointly by a working group of \emph{Plattform Industrie4.0}~\cite{plattformi40css2022}. 
A \textbf{capability} is ``an implementation-independent specification of a function in industrial production to achieve an effect in the physical or virtual world''~\cite{koecher2023reference}. 
Skills provide executable implementations of capabilities with state machines and execution interfaces; since planning only requires the abstract preconditions and effects of each capability, skill-level details are not needed here.

To make capabilities precise enough for formal reasoning, K\"ocher et al.~\cite{koecher2020capability} formalize them using established standards: VDI3682~\cite{vdi3682} provides the \emph{ProcessOperator} concept that structures each capability into inputs (required states), outputs (resulting states), and process parameters. 
IEC61360-1~\cite{iec61360} provides semantic qualifiers, in particular \texttt{ExpressionGoal} (\texttt{Requirement}, \texttt{Assurance}, \texttt{ActualValue}) and \texttt{InterpretationLogic} (\texttt{Equal}, \texttt{NotEqual}), to classify each state description. 
This level of formalization is what enables the transformation to planning problems.

The CSS model explicitly identifies planning and matchmaking as use cases for the capability level~\cite{koecher2023reference}; our contribution provides a concrete implementation of this use case through AAS-native structures.

\subsection{Related Work}

Several approaches address automated PDDL generation from structured engineering models.
Balakirsky~\cite{balakirsky2015ontology} was among the first to generate PDDL from OWL ontologies, targeting manufacturing kitting applications.
Malburg et al.~\cite{malburg2023sws2pddl} propose SWS2PDDL, which transforms OWL-S semantic web service descriptions into PDDL planning domains.
Vieira da Silva et al.~\cite{vieira2023pddl} demonstrate that capability models following K\"ocher et al.'s~\cite{koecher2020capability} formalization, when expressed as OWL ontologies, contain sufficient semantic information for PDDL generation.
Wally et al.~\cite{wally2019ral,wally2021leveraging} generate PDDL from IEC~62264 (ISA-95) process models, and Nabizada et al. ~\cite{nabizada2024pddl,nabizada2025sysml} derive PDDL from SysML system models. 
Rimani et al.~\cite{Rimani2021} propose a conceptual mapping from SysML functional architectures to hierarchical planning descriptions~(HDDL).
Yet none of these approaches operate directly on the AAS and its respective submodels.

Recent work has explored large language models for generating PDDL from natural language~\cite{tantakoun2025llm}. While syntactically capable, these methods lack integration with engineering data and traceability to formal system descriptions.

Bernhard et al.~\cite{bernhard2024aas} were the first to address AAS-to-PDDL directly, adding a dedicated planning-description submodel that encodes PDDL object types, predicates, and actions and extending skill submodels with precondition/effect structures; the automated transformation is left to future work.
Weber et al.~\cite{weber2025dlr} sketch a multi-layer concept that derives PDDL from AAS skill submodels, illustrated on a toy assembly task.
In contrast, our approach derives all planning information from existing capability standards (VDI~3682, IEC~61360-1, IDTA~02011/02016) without introducing PDDL-specific AAS structures, resolving information distributed across $n$ component AAS in a shared object store.

\section{AAS-to-PDDL Transformation}
\label{sec:approach}
\begin{figure*}[t]
  \centering
  \includegraphics[width=\textwidth]{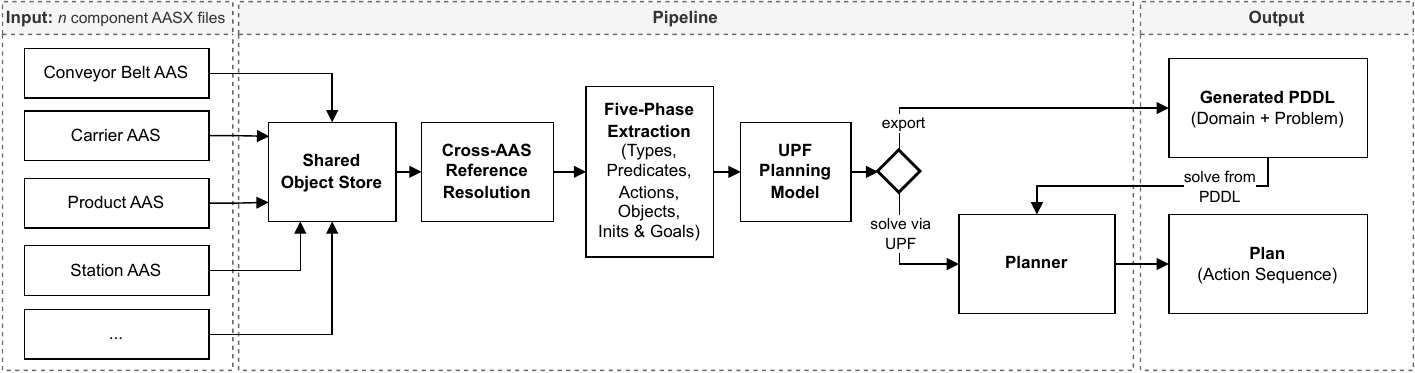}
  \caption{Overview of the proposed AAS-to-planning workflow. Multiple component AASX files are loaded into a shared object store, cross-AAS references are resolved, and planning elements are extracted in five phases. The extracted elements are assembled into a UPF planning model, which can be solved via UPF-integrated planners or exported as PDDL domain and problem files for use with any standard planner.}
  \label{fig:pipeline}
\end{figure*}

Our approach takes a set of AASX files, each describing one component of a production system, and automatically generates a complete PDDL domain and problem file. 
The key challenge is that planning-relevant information is distributed across multiple AAS files: type hierarchies, predicate definitions, capability descriptions, and instance data that reside in different components and reference each other through the AAS reference mechanism. 
Our algorithm loads all files into a shared registry, resolves these cross-references, and extracts PDDL elements in five phases (Fig.~\ref{fig:pipeline}).
The following subsections detail the underlying schema, the extraction algorithm, the reference resolution mechanism, and the implementation.

\subsection{AAS Schema and Mapping}

Our approach requires each component in the production system to be modeled in its own AAS file\footnote{Schema specification: \url{https://github.com/hsu-aut/AAS-Planning-Metamodel}. Implementation: \url{https://github.com/hsu-aut/AAS2PDDL}}, following a schema that combines the standards introduced in Section~\ref{sec:background}.
This Multi-AAS architecture enables modular composition: components can be added or removed independently, and only connectivity information needs updating.

\begin{figure}[htbp]
  \centering
  \includegraphics[width=\linewidth]{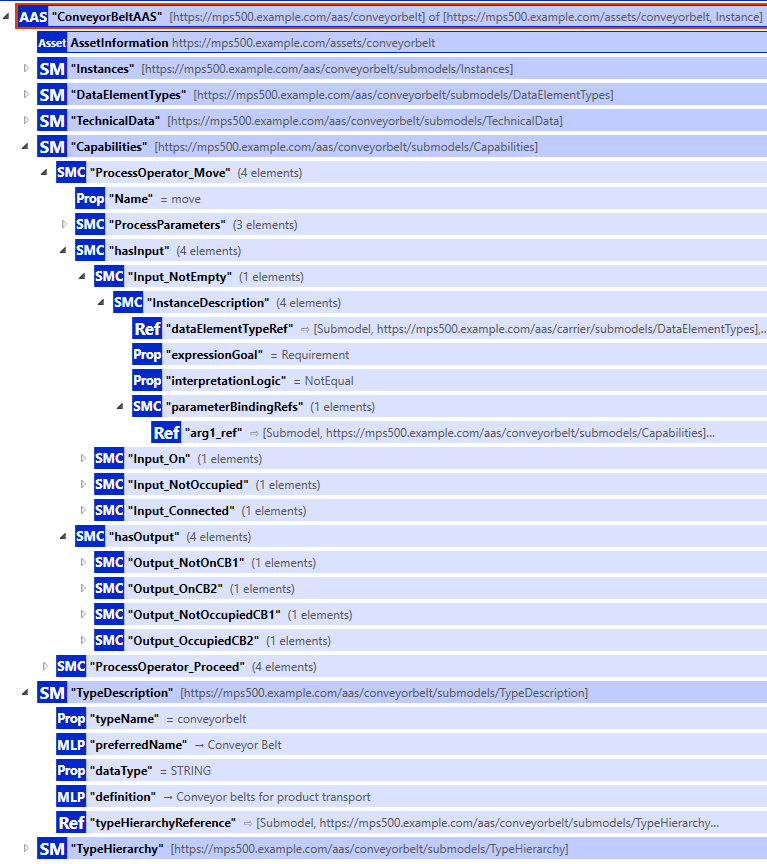}
  \caption{A component AAS in the AASX Package Explorer, illustrating the schema all component AAS follow: \texttt{TypeHierarchy} (IDTA 02011), \texttt{DataElementTypes} (IEC 61360), \texttt{Capabilities} (VDI 3682), and \texttt{Instances} (IDTA 02016), alongside \texttt{TechnicalData} and \texttt{TypeDescription}. The expanded \texttt{ProcessOperator\_Move} shows how a capability's \texttt{hasInput}/\texttt{hasOutput} states carry IEC~61360 qualifiers (\texttt{expressionGoal}, \texttt{interpretationLogic}) and reference predicate templates in other AAS via \texttt{dataElementTypeRef}. All shown element types (\texttt{SMC}, \texttt{Prop}, \texttt{Ref}, \texttt{MLP}) are generic AAS metamodel primitives, not PDDL-specific structures.}
  \label{fig:aasview}
\end{figure}

Each component AAS contains four submodels corresponding to the standards from Section~\ref{sec:background}: \emph{TypeHierarchy}, \emph{DataElementTypes}, \emph{Capabilities}, and \emph{Instances} (Fig.~\ref{fig:aasview}).
These submodels use \emph{ReferenceElements} to link to each other and across AAS boundaries; for example, a ProcessOperator's input references a DataElementType via \texttt{dataElementTypeRef}, and an instance references its type via \texttt{instanceTypeRef} (see Section~\ref{sec:crossref}).

Table~\ref{tab:mapping} shows the complete element-level mapping from AAS to PDDL.

\begin{table}[htbp]
\centering
\caption{Mapping from AAS Elements to PDDL}
\label{tab:mapping}
\resizebox{\columnwidth}{!}{
\begin{tabular}{@{}lll@{}}
\toprule
\textbf{PDDL Element} & \textbf{AAS Source} & \textbf{Standard} \\
\midrule
Type name & Entity.idShort & IDTA 02011 \\
Type parent & Entity $\rightarrow$ Parent ref & IDTA 02011 \\
Predicate name & DataElementType.idShort & IEC 61360 \\
Predicate param & DataElementType param/typeDefinitionRef & IEC 61360 \\
Action name & Capability (ProcessOperator).name & VDI 3682 \\
Action param & ProcessOperator param/typeDefinitionRef & VDI 3682 \\
Precond. (+) & hasInput, Goal=Req., Equal & VDI 3682 + IEC 61360 \\
Precond. ($-$) & hasInput, Goal=Req., NotEq & VDI 3682 + IEC 61360 \\
Effect (+) & hasOutput, Goal=Assur., Equal & VDI 3682 + IEC 61360 \\
Effect ($-$) & hasOutput, Goal=Assur., NotEq & VDI 3682 + IEC 61360 \\
Object name & instance/instanceName & IDTA 02016 \\
Object type & instanceTypeRef & IDTA 02016 \\
Init fact & initialStates, Goal=Actual & IEC 61360 \\
Goal fact & goals, Goal=Req. & IEC 61360 \\
\bottomrule
\end{tabular}}
\end{table}

To illustrate, Listing~\ref{lst:pddl} shows the generated \texttt{move} action, derived from the VDI~3682 ProcessOperator visible in Fig.~\ref{fig:aasview}; the parameters \texttt{?c}, \texttt{?cb1}, \texttt{?cb2} correspond directly to the \texttt{ProcessParameters} in the AAS.
The action moves a loaded carrier between connected conveyor belt segments.
Preconditions are derived from \texttt{hasInput} states with \texttt{ExpressionGoal=Requirement}: the carrier must be loaded (\texttt{Input\_NotEmpty}, \texttt{NotEqual}), on the source belt (\texttt{Input\_On}), the source and destination must be connected (\texttt{Input\_Connected}), and the destination must be unoccupied (\texttt{Input\_NotOccupied}).
Effects are derived from \texttt{hasOutput} with \texttt{ExpressionGoal=Assurance}: the carrier moves to the destination and occupancy changes accordingly.

\begin{figure}[htbp]
\begin{lstlisting}[style=pddl, caption={Generated PDDL action for carrier movement, derived from the AAS in Fig.~\ref{fig:aasview} (parameter suffixes formatted)}, label=lst:pddl]
(:action move
  :parameters (?c - carrier
               ?cb1 - conveyorbelt
               ?cb2 - conveyorbelt)
  :precondition (and
    (not (empty ?c))
    (on ?c ?cb1)
    (not (occupied ?cb2))
    (connected ?cb1 ?cb2))
  :effect (and
    (not (on ?c ?cb1))
    (on ?c ?cb2)
    (not (occupied ?cb1))
    (occupied ?cb2)))
\end{lstlisting}
\end{figure}

\subsection{Five-Phase Extraction Algorithm}

Algorithm~\ref{alg:extraction} presents the transformation procedure. All AASX files are first loaded into a shared object store~$\mathcal{O}$ that enables cross-AAS reference resolution (Section~\ref{sec:crossref}). The algorithm then extracts PDDL elements in five phases (cf.\ Fig.~\ref{fig:pipeline}), each targeting a specific standard.

\begin{algorithm}[htbp]
\caption{AAS-to-PDDL Transformation}
\label{alg:extraction}
\small
\begin{algorithmic}[1]
\REQUIRE Set of AASX files $\mathcal{A} = \{a_1, \ldots, a_n\}$
\ENSURE PDDL domain $D$, problem $P$
\STATE Load all $a_i$ into shared object store $\mathcal{O}$
\STATE \textbf{Phase 1} (IDTA 02011): Extract type hierarchy
\FORALL{Entity $e$ under EntryNode, recursive}
  \STATE $D$.types $\leftarrow$ $D$.types $\cup$ \{($e$.idShort, parent($e$))\}
\ENDFOR
\STATE \textbf{Phase 2} (IEC 61360): Extract predicates
\FORALL{DataElementType $d$ in DataElementTypes submodels}
  \STATE name $\leftarrow$ deriveFromIdShort($d$.idShort)
  \FORALL{parameter $p$ in $d$.parameters}
    \STATE $p$.type $\leftarrow$ resolve($p$.typeDefinitionRef, $\mathcal{O}$)
  \ENDFOR
  \STATE $D$.predicates $\leftarrow$ $D$.predicates $\cup$ \{(name, params)\}
\ENDFOR
\STATE \textbf{Phase 3} (VDI 3682): Extract actions
\FORALL{ProcessOperator $op$ in Capabilities submodels}
  \FORALL{state $s$ in $op$.hasInput $\cup$ $op$.hasOutput}
    \STATE pred $\leftarrow$ resolve($s$.dataElementTypeRef, $\mathcal{O}$)
    \STATE negate $\leftarrow$ ($s$.interpretationLogic = \texttt{NotEqual})
    \IF{$s$.expressionGoal = \texttt{Requirement}}
      \STATE pre $\leftarrow$ pre $\cup$ \{(pred, $s$.bindings, negate)\}
    \ELSIF{$s$.expressionGoal = \texttt{Assurance}}
      \STATE eff $\leftarrow$ eff $\cup$ \{(pred, $s$.bindings, negate)\}
    \ENDIF
  \ENDFOR
  \STATE $D$.actions $\leftarrow$ $D$.actions $\cup$ \{($op$.name, params, pre, eff)\}
\ENDFOR
\STATE \textbf{Phase 4} (IDTA 02016): Extract objects
\FORALL{instance $i$ in Instances submodels}
  \STATE type $\leftarrow$ resolve($i$.instanceTypeRef, $\mathcal{O}$)
  \STATE $P$.objects $\leftarrow$ $P$.objects $\cup$ \{($i$.name, type)\}
\ENDFOR
\STATE \textbf{Phase 5} (IEC 61360): Extract init and goals
\FORALL{state $s$ in instance initialStates $\cup$ goals}
  \STATE pred $\leftarrow$ resolve($s$.dataElementTypeRef, $\mathcal{O}$)
  \IF{$s$.expressionGoal = \texttt{ActualValue}}
    \STATE $P$.init $\leftarrow$ $P$.init $\cup$ \{ground(pred, $s$.bindings)\}
  \ELSIF{$s$.expressionGoal = \texttt{Requirement}}
    \STATE $P$.goal $\leftarrow$ $P$.goal $\cup$ \{ground(pred, $s$.bindings)\}
  \ENDIF
\ENDFOR
\RETURN $D$, $P$
\end{algorithmic}
\end{algorithm}

We highlight three aspects of the algorithm.

\textbf{Reference-driven extraction.}
Every phase uses \texttt{resolve(ref, $\mathcal{O}$)} to follow ReferenceElements across submodel and AAS boundaries.
No PDDL element names or type mappings are hardcoded; all information is retrieved by traversing the AAS reference graph.
This means the algorithm generalizes to any production system that follows the schema from Section~\ref{sec:approach}, without modification.

\textbf{Predicate name derivation.}
PDDL predicate names are derived from the \texttt{idShort} of DataElementType elements (e.g., \texttt{DataElementType\_On} $\rightarrow$ \texttt{on}), not from human-readable \texttt{preferredName} fields which may contain spaces.

\textbf{Encoding positive and negative conditions.}
IEC~61360-1 provides two attributes that jointly determine the PDDL semantics of each state description:
\texttt{ExpressionGoal} classifies the role within a capability description (\texttt{Requirement} = precondition, \texttt{Assurance} = effect) and within instance state descriptions (\texttt{ActualValue} = initial state, \texttt{Requirement} = goal), while \texttt{InterpretationLogic} determines polarity (\texttt{Equal} = positive, \texttt{NotEqual} = negated). 
This combination handles all four cases uniformly:
a positive precondition (\texttt{Requirement} + \texttt{Equal}), a negative precondition (\texttt{Requirement} + \texttt{NotEqual}), a positive effect (\texttt{Assurance} + \texttt{Equal}), and a negative/delete effect (\texttt{Assurance} + \texttt{NotEqual}).
For example, the \texttt{move} action (Listing~\ref{lst:pddl}) models \texttt{on(?c, ?cb2)} as a positive effect and \texttt{not(on(?c, ?cb1))} as a negative effect, both explicitly encoded through their respective \texttt{InterpretationLogic} values in the AAS data.

\subsection{Cross-AAS Reference Resolution}
\label{sec:crossref}

In the Multi-AAS architecture, planning-relevant information is distributed across files.
For example, a ProcessOperator in the ConveyorBelt AAS references predicate definitions from the Carrier AAS (e.g., the \texttt{on} predicate).
Similarly, instance definitions reference type definitions from other AAS files.

The AAS standard defines a reference mechanism based on globally unique identifiers: each AAS, submodel, and element has a unique key, with ReferenceElements pointing to these keys.
Our algorithm exploits this by loading all AASX files into a single in-memory registry (the \emph{object store} $\mathcal{O}$ in Algorithm~\ref{alg:extraction}; cf.\ Fig.~\ref{fig:pipeline}).
Since AAS identifiers are globally unique by specification, no naming conflicts arise when merging multiple files into a shared registry. 
When the extraction encounters a ReferenceElement, it resolves the target by looking up the referenced key in $\mathcal{O}$, regardless of which AAS file originally defined the target element.

This mechanism supports four types of cross-references used in our schema:
\begin{itemize}
    \item \textbf{\texttt{typeDefinitionRef}}: Links a parameter to its type definition (e.g., parameter \texttt{c} $\rightarrow$ type \texttt{carrier} from CarrierAAS).
    \item \textbf{\texttt{dataElementTypeRef}}: Links a state description to a predicate template (e.g., hasInput state $\rightarrow$ \texttt{DataElementType\_On} from ConveyorBeltAAS).
    \item \textbf{\texttt{parameterBindingRef}}: Links a state parameter to the corresponding action parameter variable.
    \item \textbf{\texttt{instanceTypeRef}}: Links an instance to its type definition, which may reside in a different AAS.
\end{itemize}

This reference-based architecture enables component-level modularity: adding a new station type requires only creating a new AAS file with references to shared type and predicate definitions. 
Instance-level integration such as connectivity between conveyor segments is modeled as initial state facts using IEC~61360-1 expressions within the component AAS. Layout changes thus require editing these instance definitions.

\subsection{Implementation}

The transformation is implemented using the BaSyx SDK~\cite{basyxsdk} for AAS parsing and reference resolution, and UPF~\cite{upf2025} for constructing the planning problem programmatically (Fig.~\ref{fig:pipeline}).
UPF provides a planner-independent API: the extracted types, predicates, actions, and problem elements are assembled into a UPF problem object, which can then be exported to standard PDDL files or solved directly through any UPF-integrated planner.
In our case study, we use Fast Downward~\cite{helmert2006fast} via UPF~1.3.0 with the \texttt{MinimizeSequentialPlanLength} quality metric to obtain shortest plans; switching to a different planner requires only a configuration change.
The only input required by our toolchain are the AAS files themselves; the implementation and all example AAS models used in this paper are publicly available\footnotemark[1] to support reproducibility.

\section{Case Study}
\label{sec:evaluation}

\subsection{Festo MPS500 Production System}

We apply our approach to the Festo MPS~500, a modular production system used in automation education and research. 
The case study is performed entirely on the AAS models; no physical system is involved. 
The system consists of 30 conveyor belt segments in a circular layout with inner shortcuts, seven stations (raw material storages, processing, quality check, robotic assembly, buffer storage, shipping), four carriers, and two product types.

\begin{figure}[t]
    \centering
    \includegraphics[width=0.84\linewidth]{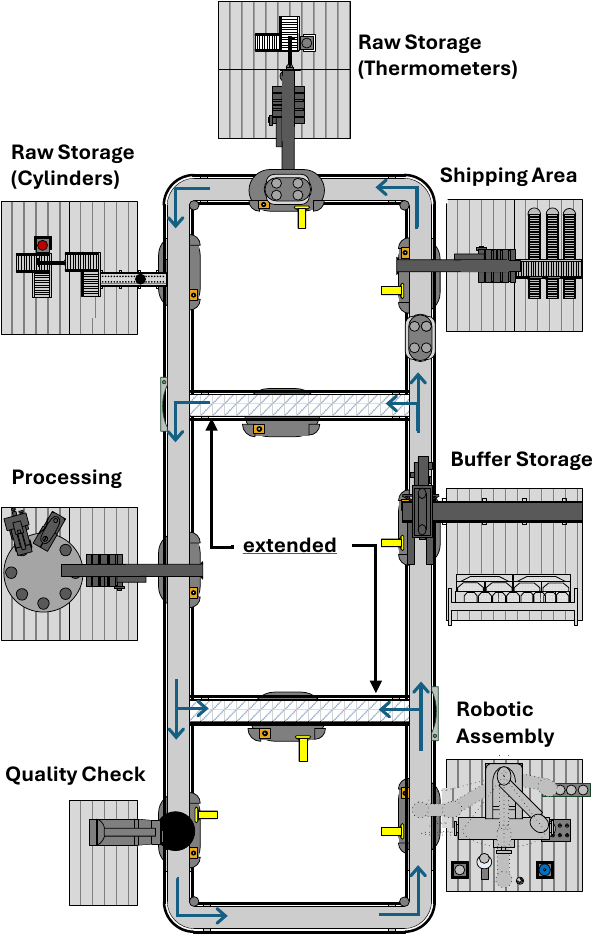}
    \caption{Schematic layout of the Festo MPS500. Arrows indicate conveyor directions; the two hatched inner paths (upper, lower) provide bypass routes evaluated in the design space exploration.}
    \label{fig:mps500}
\end{figure}

Following the workflow in Fig.~\ref{fig:pipeline}, the system is modeled using nine component AAS files (one per component type: six station types, conveyor belt, carrier, and product), yielding 11 PDDL types (including two abstract supertypes), 11 predicates, 9 actions, and 45 objects.
Figure~\ref{fig:mps500} shows the schematic layout. 
The generated PDDL files can be solved by any standard planner, enabling automated plan generation directly from AAS models.
The pipeline generates these files in about 0.3\,s on an AMD Ryzen~7~PRO~8840HS with 60\,GB RAM (median of five runs), and Fast Downward accepts the output without modification, solving it to certified optimality (\texttt{SOLVED\_OPTIMALLY}); no hand-editing of the generated PDDL is required.

\subsection{Design Space Exploration}

The primary contribution of our approach is that engineers can generate complete planning problems directly from their AAS models without PDDL expertise. 
To illustrate the practical utility beyond plan generation itself, we demonstrate a design space exploration across four layout variants by modifying the inner conveyor paths shown in Figure~\ref{fig:mps500}. 
All conveyor segments are unidirectional; the outer ring moves counterclockwise.
The two inner paths provide shortcuts between the processing side and the buffer side of the ring, each carrying traffic in one direction (upper path upward, lower path downward).
The four variants differ in which combination of inner paths is available (Table~\ref{tab:dse}).

The production scenario requires shipping one thermometer product. Starting from the raw material warehouse, the product must be loaded onto a carrier, transported through processing, quality control, robotic assembly, and finally shipped.
The product carrier always travels the outer ring through all stations.
The inner paths do not enable skipping stations.
Instead, the inner paths serve as bypass routes for \emph{idle carriers}: when the product carrier needs to pass through a segment occupied by another carrier, the idle carrier must move out of the way.
The inner paths provide alternative routes for this traffic management.
Each variant is represented by its own set of AAS files, differing only in the \texttt{connected} predicates of the ConveyorBelt AAS. The PDDL files are regenerated from each variant's AAS directory, and optimal plans are computed. Table~\ref{tab:dse} summarizes the results.

\begin{table}[htbp]
\centering
\caption{Design Space Exploration Results (Optimal Planning)}
\label{tab:dse}
\begin{tabular}{lccc}
\toprule
\textbf{Variant} & \textbf{Inner Paths} & \textbf{Plan Length} & \textbf{Savings} \\
\midrule
A (Full) & upper + lower & 35 steps & 46\% \\
D (Lower only) & lower & 35 steps & 46\% \\
C (Upper only) & upper & 59 steps & 9\% \\
B (No inner) & none & 65 steps & --- \\
\bottomrule
\end{tabular}
\end{table}

Variants~A and~D yield identical 35-step optimal plans, indicating that only the lower inner path is used for traffic management in this scenario.
Without this path (Variants~B and~C), idle carriers must be shuffled along the outer ring, increasing plan length to 65 and 59 steps respectively.
The upper inner path alone (Variant~C) provides modest savings (9\%) by offering an alternative return route for displaced carriers.

The planner reveals that inner paths primarily reduce \emph{traffic congestion} rather than shortening the product's route, a non-obvious interaction between topology and multi-carrier logistics that is difficult to anticipate manually.
Engineers can explore such trade-offs by modifying the AAS model and regenerating the planning domain, without requiring PDDL expertise.
Note that a comprehensive exploration would additionally vary initial states and goal conditions across scenarios; here we fix the production scenario to isolate the effect of topology changes.

Across the four variants, planner search time ranges from 6.8\,s (variant~D) to 219\,s (variant~B), while PDDL generation remains a sub-second, model-independent overhead.
The variants are obtained entirely by edits at the AAS level --- specifically, to the \texttt{connected} facts of the ConveyorBelt AAS --- yielding roughly 520 lines of automatically generated PDDL across the four runs that the engineer never writes.

\section{Conclusion}
\label{sec:conclusion}
This paper presents an approach for automatic PDDL generation from AAS capability models, addressing two research questions. 
Regarding RQ1~(how to systematically transform AAS capability models into PDDL), we showed that AAS capability models structured using VDI3682, IEC61360-1, and IDTA~02011/02016 contain sufficient information to derive all PDDL elements, without introducing PDDL-specific AAS structures. 
Regarding RQ2~ (how to handle distributed Multi-AAS architectures), we addressed the distribution of planning information across multiple AAS files through a shared object store that resolves cross-AAS references, enabling a five-phase extraction algorithm to operate across distributed Multi-AAS architectures. 
Since the output is standard PDDL, any compliant planner can be used, and the UPF integration makes switching planners a mere configuration change.

The case study on a Festo MPS~500 demonstrated the possibility of design space exploration across four layout variants, revealing that inner conveyor paths reduce optimal plan length through improved carrier traffic management. Engineers can explore such trade-offs by modifying AAS files and regenerating the planning domain, without requiring PDDL expertise.

Current limitations include the restriction to PDDL with typing and negative preconditions. 
Although VDI~3682 and IEC~61360 also support duration and numeric data elements, mapping them to PDDL~2.1~\cite{fox2003pddl21} durative actions and numeric fluents is left to future work. 
In this work, the component AAS are modeled manually; approaches exist to derive such capability models (semi-)automatically, e.g., from Module Type Package (MTP) descriptions~\cite{koecher2022mtp} or using large language models~\cite{vieira2024llm}, which could be combined with our method to further reduce the modeling effort. 
Validating the approach on industrial-scale systems is an important next step. 
Integration with MBSE approaches for early-phase production system design~\cite{beers2026dsml} could enable a continuous model chain from conceptual design models through AAS to PDDL.

\bibliographystyle{IEEEtran}
\bibliography{references}

\end{document}